\documentclass[a4paper,conference]{IEEEtran}
\ifCLASSINFOpdf
 \usepackage[pdftex]{graphicx}
\else
\fi
\usepackage{amsmath}

\begin{document}
\title{Deep Feature Fusion for Mitosis Counting}
\author{\IEEEauthorblockN{Robin E Yancey}
\IEEEauthorblockA{Department of\\Computer Science\\
University of California, Davis\\
Davis, California 95616\\
Email: reyancey@ucdavis.edu}}

\maketitle

\begin{abstract}

Each woman living in the United States has about 1 in 8 chance of developing invasive breast cancer. The mitotic cell count is one of the most common tests to assess the aggressiveness or grade of breast cancer. In this prognosis, histopathology images must be examined by a pathologist using high-resolution microscopes to count the cells. Unfortunately, this can be an exhaustive task with poor reproducibility, especially for non-experts. Deep learning networks have recently been adapted to medical applications which are able to automatically localize these regions of interest. However, these region-based networks lack the ability to take advantage of the segmentation features produced by a full image CNN which are often used as a sole method of detection. Therefore, the proposed method leverages Faster RCNN for object detection while fusing segmentation features generated by a UNet with RGB image features to achieve an F-score of 0.508 on the MITOS-ATYPIA 2014 mitosis counting challenge dataset, outperforming state-of-the-art methods.

\end{abstract}


\IEEEpeerreviewmaketitle

\section{Introduction}

\subsection{Mitotic Count \& Issues}

It is estimated that breast cancer incidences will increase by more than 50\% by 2030 from 2011. And, a technique called \textit{mitosis counting} is one of the most commonly used methods of assessing the level of progression. It is also a routine task for every patient diagnosed with invasive cancer. 

The Nottingham Grading System (NGS) is recommended by the World Health Organization [WHO], American Joint Committee on Cancer [AJCC], European Union [EU], and the Royal College of Pathologists (UK RCPath) \cite{ngs2}. It says that final score for tubule formation, nuclear pleomorphism, and mitotic index should be considered on a scale from 3 and 9. These are then categorized into grades where Grade 1 is score 3–5, or well differentiated; Grade 2 is score 6–7, or moderately differentiated; and Grade 3 is score 8–9, or poorly differentiated \cite{NGS}. 

Pathologists assessments of the progression of cancer are based on the mitotic count for each tumor. This is defined on a region found with high proliferative activity (or the highest mitotic count) where 10 High-Power Fields are obtained within a space of $2mm^{2}$.

Although mitotic count is the strongest prognostic value, it is a tedious and subjective task with poor reproducibility, especially for non-experts. Pathologists are highly trained the process of obtaining the slides, but using different scanners and different preparation techniques may make distinguishing cells more tiring and difficult. 

What makes matters more complicated is that mitotic cells differ based on whether the mitosis is in one of the four main phases: prophase, metaphase, anaphase, and telophase. In telopahse for example, cells are split into two parts but still barely connected. Further, apoptotic cells going through cell-death can resemble mitosis \cite{tupac}. Since biopsies can take up to ten days before the patient receives results, and cancer incidences continuing to rise. \cite{1} a more accessible way a better way to provide this assessment is in need. Fortunately, inter-observer agreement with human pathologists has recently been observed when compared to automatic methods \cite{9}.

\section{Related Work}

\subsection{Automatic and Machine Learning Methods}

The use and development of automatic detection methods of mitosis counting have gradually been increasing since the end of the 20th century in order to make doctors' jobs easier and more efficient \cite{mitosisreview}. Due to the recent progress in digital medication, a large amount of of data has became available for use in the medical studies. Machine learning has helped to discover new characteristics of cancer mutations by sorting through more image data than humanly possible and simultaneously analyzing all of the millions of image pixels undetectable to the human eye. For example, in the field of histopathology, machine learning algorithms have been used for analysis of scanned slides to assist in tasks including diagnosis \cite{5}. The use of computing in image analysis may reduce variability in interpretation, improve classification accuracy, and provide clinicians (or those in training) with a second opinion \cite{5}. Existing methods use either handcrafted features captured by specific morphological, statistical, or textural attributes determined by a pathologist or features are automatically learned through the use of convolutional neural networks (CNN).

\subsection{Hand Carved Feature Extraction} 
Techniques used to count mitosis in HPF images typically fall into the category of those based on handcrafted features or CNN features. Early automatic detection methods of mitosis mainly consisted of machine recognition based off of hand-carved or manually designed features. Manually designed features are required to be described by a human to represent the characteristics of mitosis. This is usually based on a human's professional domain knowledge of appropriate mitosis textural or morphological features. On the other hand, with newer deep learning, methods these features are able to be discovered by the machine, such as with the use of neural networks. 

 Current methods also often fall into the category of considering mitosis detection as a classification problem or as a semantic segmentation problem. When it is framed as a classification problem each of the regions are usually classified as containing a mitosis or no-mitosis using a sliding window across the image. On the other hand, when it is framed as a segmentation problem a fully connected network is often used to give each pixel in the image a given intensity where a threshold determines whether or not it is a pixel contained in a mitosis. 
For example, \cite{6620101} first used MLE (Maxiumum Likelihood Estimation) to extract pixel-wise features based on color information captured by MLE. Then they distinguished mitotic regions from non-mitotic regions using the extracted textural features obtained from CLBP (Completed Local Binary Pattern). And, lastly they used an SVM (Support Vector Machine) classifier to classify extracted feature vectors. \cite{7165640} used the maximization of relative-entropy between the cells and the background to segment and then used a RF classified to discriminate mitotic from non-mitotic regions. Zhang et al. proposed a method that fuses handcrafted features based on four measure indices of the GLCM (Gray Level Co-occurrence Matrix) and deep features based on natural image knowledge transfer to segment mitotic cells \cite{rev1}. Irshad first analyzed statistics and morphological features in specific channels of various color spaces which are used to assist pathologists in mitosis extraction. Next, they then performed a LoG (Laplacian of Gaussian), thresholding, morphology, and active contour model on blue-ratio image to find and segment mitotic candidates. Candidates are selected based on first and second order morphological and textural features to be classified using a DT (Decision Tree) classifier \cite{rev2}. 
%


(Liu et al., 2010) proposed a two part mitosis event detection technique using hidden conditional random field, where candidate spatiotemporal sub-regions were segmented by image pre-conditioning and volumetric segmentation \cite{liu}. (Huh et al. 2011), use phase-contrast microscopy images consisting of three steps: candidate patch sequence construction, visual feature extraction, and identification of mitosis occurrence/temporal localization of birth event \cite{huh}. (Khan et al., 2012) modeled mitosis intensities by a gamma distribution and those from non-mitotic regions by a gaussian distribution \cite{ggmm}. (Sommer et al., 2012) trained a pixel-wise classifier is to segment candidate cells, which are classified into mitotic and non-mitotic cells based on object shape and texture \cite{sommer}. (Huang \& Lee, 2012) propose an extension of a generic ICA, focusing the components of differences between the two classes of training patterns \cite{huang}. (Veta et al., 2013) extracts image segmentation with the Chan-Vese level set method and classifies potential objects based on a number of size, shape, color and texture features \cite{veta}. (Tek et al. 2013) use an ensemble of cascade adaboosts classified mainly by a shape-based feature, which counted granularity and red, green, and blue channel statistics \cite{tek}. 

There have been a few groups using handcrafted-features to report F-measure scores close to the highest score achieved in the ICPR 2012 contest. For example, (Irshad, 2013) proposed a method where cells are segmented by extracting various color channel features followed by a candidate detection that included Laplacian of Gaussian, thresholding, and morphological operations, achieving an F-measure score of 0.72 on the ICPR 2012 dataset. (Irshad et al., 2013) then extended this model combining texture features with decision tree and SVM classifiers increasing to an F-measure score of 0.76. (Nateghi et al., 2014) omit non-mitosis candidates by a cast function and by minimization using Genetic Optimization, before co-occurrence, run-length matrices and Gabor features are extracted from the rest of candidates to classify mitosis with SVM, achieving an F-score of 0.78 \cite{nateghi}.

\subsection{Deep Learning Methods} 

\subsubsection{Combined Methods}
Although handcrafted features are domain-inspired for the particular application, data-driven CNN models have the ability to learn additional feature bases that cannot be represented through any of the handcrafted features. Some former methods also have combined machine learning with hand-carved features in order to exploit the advantages of hand generated features and combine them with CNN. \cite{combined}. These usually use two main steps: segmentation and classification \cite{mitosisreview}. In the first step candidate regions are segmented to decrease the size of the input image locations that will need to be tested (usually by using maximum likelihood estimation, threshold segmentation, and watershed segmentation). In the second step the classification algorithm is trained specifically to detect the hand described features to extract mitosis from non-mitosis in the pre-segmented region. 

For example, (Wang et al., 2014) present a cascaded approach separately extracting each type of features for each sub-image region, achieving an F-score of 0.73 on ICPR 2012. Regions where the two individual classifiers disagree are further classified by a third classifier, and the final prediction score is a weighted average of the outputs of all classifiers \cite{combine}. (Malon et al., 2013) combines color, texture, and shape features from segmented mitosis regions with extracted by convolutional neural networks (CNN), achieving an 0.66 F-score \cite{malon}.

\subsubsection{CNN Methods}
Most of the latest most-efficient machine algorithms for mitosis detection have been deep learning-based since it is hard to manually design meaningful and discriminative features well enough to separate a mitosis from a non-mitosis due to the variation and complexity of the shapes. With a large number of labeled training images feature extraction is left to the DNN which is optimized using mini-batch gradient descent. Low level layers learn the low level features of the training images, while high level features are learned by the deep layers. Although data-driven models are more computationally expensive, they have the advantage of being able to detect features that cannot be represented by handcrafted features. 

Mitotic figures are ideal for CNNs with their high level features and textures. In fact, CNN methods were used to win the ICPR2012 \cite{icpr12}, ICPR 2014 \cite{icpr14}, and AMIDA 2013 challenges \cite{amida13} \cite{1}. These are three well-known mitosis counting competitions and were held at conferences. The datasets are publicly available and commonly used for research \ref{sssec:DS2}. For example, (Cireşan et al., 2013) used a feed-forward net made of successive pairs of convolutional and max-pooling layers, followed by several fully connected layers \cite{ciresan}. This approach achieved an F-measure score of 0.78, the winning score in the ICPR 2012 contest summarized in \cite{icpr12}, yet it requires multiple days for training and making it not clinically applicable. 

Additionally, (Albayrak et al., 2016) developed a CNN which uses a combination of PCA and LDA dimension reduction methods for regularization and dimension reduction process before classification of mitosis by an SVM \cite{Albayrak}. (Chen et al., 2016) propose a method combining a deep cascaded CNN to locate mitosis candidates and a discrimination model using transferred cross-domain knowledge, achieving an F1 score of 0.79 on ICPR 2012 \cite{chen}. (Wu et al., 2017) used deep fused fully convolutional neural network modified to combine low and high level layer together using to reduce over-fitting \cite{wu}. (Xue \& Ray, 2018) train a CNN to predict and encode a compressed vector of fixed dimension formed by random projections of the output pixels, achieving state-of-the-art results \cite{xue}. (Wahab et al., 2019) used a pre-trained CNN for segmentation and then a Hybrid-CNN for mitosis classification, achieving an F-score of 0.71 on ICPR 2012 \cite{wahab}.

\subsubsection{Issues with Non-regional CNN-based Methods}
Although CNNs have high classification accuracy, they can also have a high computational load. Further, since background tissue regions which are mitotic and non-mitotic can be hard to differentiate, pre-processing the full input image prior to applying the CNN can help reduce the work. Therefore, a number of groups have tried to alleviate this issue by designing course candidate extractors for mitotic regions prior to input into the CNN. For example, Yuguang Li et al. used a feature-based region extractor plus two CNN stages to achieve an F-score of 0.78 on the ICPR 2012 dataset and an F-score of 0.43 on the ICPR 2014 dataset \cite{li}. As another example, Chen et al. used a course retrieval model prior to a three layer FCN (Fully Connected Network) followed by a CaffeNet \cite{caffe} to classify patches and achieved an F-score of 0.79 on the ICPR 2012 dataset \cite{Chen_Dou_Wang_Qin_Heng_2016}.

\subsection{Background \& Motivation} 

Use of CNN for this application will require limiting the size of the input image in order for features to be extracted from sub-image patches. This is in order to localize features within regions of the image rather than the full image consisting of multiple objects or non-objects. Moreover, object detection or precise localization is actually a more common task than full-image classification in medical applications. 

Consequently, a number of region-based CNN approaches have been proposed, achieving scores which exceed those of standard deep learning. (Rao, 2018) uses a region based convolutional neural network (RCNN) which beat all three contest winners scores when trained on all three datasets from each contest \cite{rao}. (Chao Li et al., 2018) uses multi-stage deep learning based on Faster RCNN in combination with a verification model using residual networks to distinguish false positives from detections, achieving an F-score of 0.83 on the 2012 MITOSIS dataset and 0.44 on the 2014 dataset. A separate network component is included to estimate the region of mitosis when weak labels are given for training \cite{4}.

\subsubsection{RCNN}


Since the number of objects of interest is not fixed, a standard CNN followed by a fully connected layer is inadequate because objects (or mitotic figures) have different sizes and locations. It is too computationally expensive to check all of the sizes and locations. However, these RCNN models have the ability to use region proposal algorithms to extract regions of interest using selective search, SVM, and RoI pooling techniques to choose ideal candidates rather than checking all the locations. The resulting bounding boxes can then be updated on each iteration to get closer to the precise location of the ground truth. Since only the parts of the image that are likely to contain objects must be checked, this usually results in faster training times over older methods. 

This is likely why most top-performing models are RCNN-based. CasNN \cite{chen} used a coarse retrieval model to identify and localize mitotic candidates while preserving a high sensitivity. Unfortunately, this method is too slow to be clinically applicable, having an inference speed of 4.62 seconds per HPF. 

DeepMitosis \cite{4} took advantage of an RCNN model with an RPN and obtained a score of 0.437. The Lightweight RCNN \cite{li} achieves a slightly lower score of 0.427 than both which may due to the fact that it uses a course candidate extractor instead of an RPN. These methods are faster though, as the lightweight RCNN takes 0.83, and DeepMitosis takes 0.72 seconds.

\subsubsection{Faster RCNN}

Faster-RCNN \cite{faster-rcnn} is an object detection network that combines a CNN with a region proposal network (RPN) to classify sub-image region proposals. It can then refine these to output spatial coordinates for bounding boxes associated with certain classes. This model improves upon the RCNN by using the RPN that takes the feature maps of the last convolutional layer as input and outputs region proposals, making it quicker. 

\subsubsection{Multi-stream Faster RCNN}

Previously, the author applied this network to non-objects with a multi-stream version of Faster RCNN to detect image tampering \cite{10} \cite{11}. This largely outperformed the results from individual stream as well as many current 'state-of-the-art' methods on popular image manipulation datasets. The purpose of the RGB (red, green, blue color scale) stream is to find regions of high color contrast difference, while the purpose of the second stream is to extract regions with JPEG artifacts. Region proposals are taken from the RGB stream alone, but features from both streams are fused in the bilinear pooling layer to maintain the spatial concurrence of each. 

Interestingly, the classification models used for image tampering detection are similar to that of mitosis detection in HPFs in that their purpose is to localize regions of interest nearly indistinguishable to the human eye. Histopathological image segmentation is also commonly used to extract and highlight objects of interest from the background of the image (with different textures) for further identification. Some of the segmentation methods used for mitosis counting or cancer cell identification have included thresholding (using Fourier descriptors, wavelets or Otsu), edge detection (using Canny and Sobel filters), or clustering (such as k-means, mean-shift, and K-nn) \cite{7} \cite{8}. Since the segmented image map significantly improved classification accuracy in image tampering, it also can improve mitosis detection accuracy. In this work, mitosis classification accuracy is improved by applying a segmented image map generated by UNet \cite{unet} as input to a second stream of the model.

\section{Materials \& Methods}

The standard version of Faster RCNN lacks pixel to pixel alignment for a clear-cut localization of objects since the RoI Pooling layer performs coarse spatial quantization for feature extraction. This technique fails to preserve the exact spatial locations of objects and could lead to inaccuracies. The UNet \cite{unet} extension is added to the second stream of the Multi-stream Faster RCNN (MS-FRCNN) in order to improve prediction accuracy with little addition to overhead computation.

UNet is used to extrapolate the missing context by mirroring the input image in order to predict the pixels in the border regions of the image. The use of a weighted loss helps separate touching objects in cell segmentation and the use of a large number of up-sampling feature channels allows information to propagate image context to higher layers. Therefore, in this work, the segmentation features from UNet are fused with the RGB image features in a Multi-stream version of Faster RCNN for a more precise localization of the mitotic cells.

\begin{figure}[htbp]
\centerline{\includegraphics[scale=0.23]{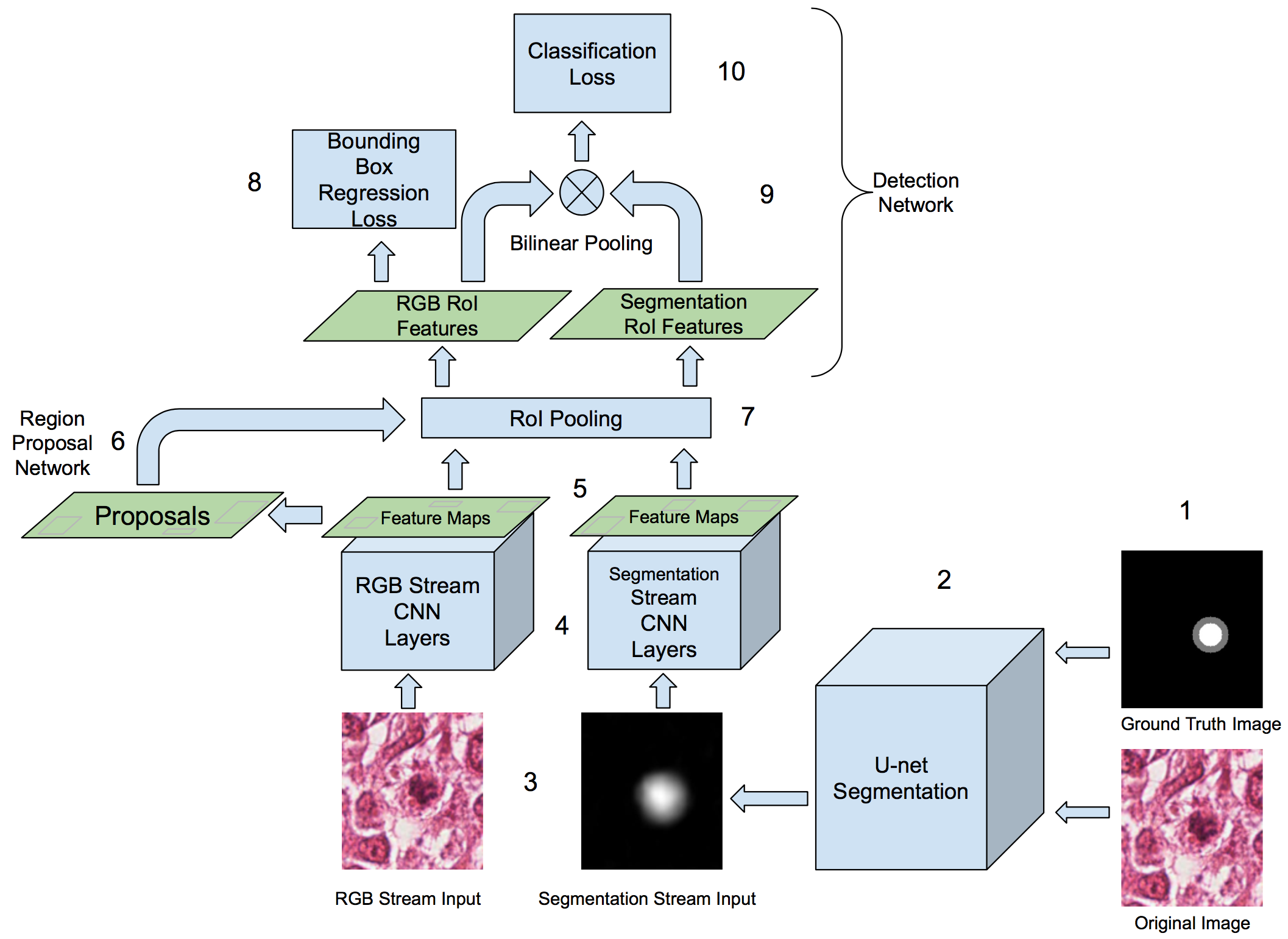}}
\caption{Multi-Stream Faster R-CNN Training Setup. 1. UNet is pre-trained using half of the RGB images along with the ground truth masks. 2. Segmented images are then produced by testing the complete set of Multi-stream Faster R-CNN training images on UNet. 3-4. Cropped RGB images are input to the RGB stream CNN, while segmented images are input to the segmentation stream input. 5. The last layer of the CNN outputs anchors with multiple scales and aspect ratios which are used for the RPN to propose regions. 6. Only the last layer of the RGB stream CNN is used as input to the RPN, so both streams share these region proposals. 7. The ROI pooling layer selects spatial features from each stream and outputs a fixed length feature vector for each proposal. 8. The RoI features from the RGB stream alone are used for the final mitosis bounding box location prediction. 9. Bilinear Pooling is used to obtain spatial co-occurrence features from both streams. 10. The final predicted classes are output from the FCN and soft-max layers using sparse cross entropy loss \cite{2}. }
\label{mfrcnn}
\end{figure}

\subsection{UNet Segmentation}
Part of the process of preparing the data for training Faster R-CNN includes applying segmentation to the images for application to the segmentation stream. This was done with a UNet \cite{unet}, as shown in the parts 1 and 2 of the block diagram of Figure \ref{mfrcnn} which lays out the training process. This a 23-layer network, where down-sampling side of the ‘U’ doubles the number of features in each step while the up-sampling side halves features in each step. The contracting path uses repeated pairs of $3 \times 3$ un-padded convolutions to capture image context before a rectified linear unit (ReLU), followed by a $2 \times 2$ max pooling operation with stride two. The symmetric expanding path helps to achieve localization and uses a $2 \times 2$ convolution and two 3x3 convolutions each followed by a ReLU. 

This image segmentation network requires ground truth images for training. This mean that a black image with a white mask in the area of localization needs to be fed into the network, for each of first the each RGB (red, green, blue color scale) image subsection (described in 5.1). This is shown in Figure \ref{mfrcnn} and is labeled part 1. Ground truth data for the segmentation network is created for each image subsection in the training set by drawing a square of black pixels the same size as each image subsection, with a white circle of pixels in the same location of mitosis using the annotated coordinates of the mitosis location. Images input into UNet must also be slightly downsized to avoid padding issues. UNet is then trained for 80 iterations on half of the Faster R-CNN training set (consisting of only the Aperio scanner images and their corresponding ground truth images), before it is tested on the images from both scanners to create the full set of segmented training data for the segmentation stream input of Faster R-CNN. It is important that at least half of the test data for UNet is not in the training set since the actual test images for the full Multi-stream network will not be in the training set of UNet. This avoids over-fitting and allows for imperfections to occur in segmented the training data.

\subsection{Multi-Stream CNN}
The framework presented in parts labeled 3 through 10 of Figure \ref{mfrcnn} is a Multi-stream version of the Faster R-CNN object detection network \cite{faster-rcnn}.  In this work, the network is adapted to classify regions of interest having a mitosis, instead of classifying actual physical objects. One stream takes an input of the segmented image map while the the other takes an input of the original RGB image. In part 3, the segmented training data output from UNet is used for input to the segmentation stream of the Multi-stream Faster-RCNN. Original images with all three color channels are used again in Faster R-CNN as input to the RGB stream. The purpose of the segmentation stream is to provide additional evidence of mitosis locations by emphasizing the segmented features picked up by UNet. 

The base convolutional neural network (CNN) in part 4 is a VGG-16 network pre-trained on ImageNet \cite{imagenet}. Nine boxes with different scales and aspect ratios are generated from each position (anchor) of the output feature map produced in the last layer of the CNN (part 5). This produces thousands of boxes that can be used for region proposals generated by the Region Proposal Network (RPN) during training (part 6).

\subsection{Region Proposal Network}
The purpose of the RPN is to select a fixed number of boxes before sending them to the Detection Network (section 4.5) to further reduce and refine them. It is a simple 3 layer convolutional network which has one layer that is fed into two sibling fully connected layers used for classification and bounding box regression, respectively. The RPN uses all the anchors selected for the mini batch to calculate the classification loss using binary cross entropy. During training, proposals that overlap the ground-truth objects with an Intersection over Union (IoU) greater than 0.5 are considered foreground, while those with a 0.1 IoU or less are considered background objects, while the rest are ignored. Bounding box regression is done using the distance vector to the nearest foreground object and uses smooth $L_1$ loss, as it is more sensitive to outliers \cite{6}.  The difference of smooth $L_1$ from $L_2$ and normal $L_1$ is that when the magnitude of the $L_1$ error is less than 1, it is considered almost correct, so it diminishes at a faster rate. 

Each of the proposals is output as this class probability representing the \textit{objectness} score paired with a more refined estimation of its rectangular coordinates \cite{Konig}. The total loss for the RPN network is defined as in \eqref{eq1} \cite{faster-rcnn}.

\begin{equation}
  \begin{array}{l}
L_{RPN} (g_i, f_i) = \frac{1}{N_{cls}} \sum_{i} L_{cls} (g_i, g_{i}^{*}) + \\
\\
\lambda \frac{1}{N_{reg}} \sum_{i} g_{i}^{*} L_{reg} (f_i, f_{i}^{*}) \label{eq1}
\end{array}
\end{equation}

 \begin{itemize}
\item $g_i$ represents the probability of anchor $i$ being a mitosis in a mini batch, 
\item $g_{i}^{*}$ is the ground-truth label for anchor $i$ being a mitosis
\item $f_i$, $f_{i}^{*}$ are the four dimensional bounding box coordinates for anchor $i$ and the ground-truth, respectively
\item $L_{cls}$ denotes cross entropy loss for RPN network, and $L_{reg}$ denotes smooth $L_1$ loss for regression
\item $N_{cls}$ denotes the size of a mini-batch, and $N_{reg}$ is the number of anchor locations.
\item $\lambda$ is a hyper-parameter to balance the two losses
 \end{itemize}

Note, that the arrow pointing towards the RPN in Figure \ref{mfrcnn} is directed from the last-layer feature map in the Faster R-CNN RGB stream, because the RPN shares convolutional features with this CNN \textit{only}, as it was found to be better than segmentation features alone or combined with RGB for measuring color changes \cite{2}, which best define borders of mitosis regions. 

\subsubsection{Non-maximum suppression}
Non-maximum suppression (NMS) is used to ensure there is not redundancy in the proposed regions, retaining only the overlapping boxes with the highest probabilities. Proposals stored in array where each RoI is defined by a four-tuple (r, c, h, w) that specifies its top-left corner (r, c) and its height and width (h, w) \cite{6}, are then sorted by score. To balance computation time we retain the top N, which is large enough so that plenty will still be classified as background.

\subsection{Region of Interest Pooling}

The Region of Interest (RoI) pooling layer (part 7) takes all of the coordinates provided by the RPN and extracts a fixed sized feature map from the the CNN using max-pooling \cite{sermanet}. The saves computation time because using the classifier on the full N proposals would be very slow. With a fixed sized feature map the detection network can be used to classify a fixed number of object classes. The segmentation and RGB streams share the RoI pooling layer, so each of these two CNN feature maps are pooled according to the coordinates from the RGB RoI. 

\subsection{Detection Network}
The Detection Network (also known as R-CNN) uses two fully connected layers which do the final classification of whether the region contains a mitosis and regress bounding boxes based on object class, respectively. Regions that do not overlap the ground truth are now ignored during training, while those that have an IoU greater than 0.5 are assigned to the mitosis class. 

\subsubsection{Bounding Box Regression}
Only the RGB stream RoI features are used to finalize object coordinates because they perform better in the RPN, as described above \cite{2}.  The offset of the bounding box to that ground truth is used to better adjust the bounding boxes, using smooth $L_1$ loss again as shown in part 8.

\subsubsection{Bilinear Pooling}
Compact bilinear pooling is used in part 9 of Figure \ref{mfrcnn} to fuse the co-occurence of features from both streams while maintaining the spatial location information \cite{gao}. Second order statistics (similar to the covariance matrix) are captured by computing the outer product of each spatial coordinate of the feature map obtained from the last layer of CNN for each stream. The output is equal to $ f_{RGB}^{T}$ $f_{seg}$, where $f_{RGB}$ is the RoI of the RGB stream and $f_{seg}$ is the RoI of the segmentation stream, and then sum-pooling is used over the feature map \cite{kong}. Finally, signed square root and $L_2$ normalization is applied before using a fully connected layer \cite{2}.

\subsubsection{Classification}

Multi-class sparse cross entropy loss is used for classification. The final predicted classes are output from the network's fully connected and soft-max layers (part 10) \cite{6}. 

\subsection{Total Loss Function}
The backward propagated signals from the RPN loss and the Fast R-CNN loss are combined for shared layers \cite{6}, so the total network loss function is the sum of all 3 of the loss components as shown in \ref{eq2}:

\begin{equation}
L_{total} = \\
L_{rpn} + L_{mit} (f_{rgb}, f_{seg}) + L_{bbox}(f_{rgb})  \label{eq2}
\end{equation}

where

\begin{itemize}
\item $L_{total}$ denotes total loss
\item $L_{rpn}$ denotes the RPN loss 
\item $L_{mit}$ denotes the final cross entropy classification loss (based on the output of bilinear pooling) 
\item $L_{bbox}$ denotes the final bounding box regression loss
\item $f_{rgb}$ represents the RoI from the RGB stream
\item $f_{seg}$ represents the RoI from the segmentation stream 
\end{itemize}

The classification and regression loss components are calculated the in the same way as in the RPN loss defined above. For example, the term $L_{mitosis} (f_{RGB}, f_{seg})$ represents the loss $\frac{1}{N_{cls}} \sum_{i} L_{cls} (g_i, g_{i}^{*})$ based on the probability $g_i$ of a ground truth obtained on the output of bilinear pooling, while $L_{bbox}(f_{RGB})$ is the regression loss based on only the coordinates from the RGB stream.

\subsection{Dataset \& Preparation} \label{sssec:DS2}

\subsubsection{Contest Datasets}

The models were trained on an open mitosis counting dataset from the International Conference on Pattern Recognition (ICPR) of breast cancer histopathology in 2014 \cite{icpr14}. \textit{Note, that this is known as a more challenging extension of ICPR 2012.} Image data is stained with standard hematoxylin and eosin (H\&E) dyes obtained from breast biopsies. The Aperio Scanscope XT and the Hamamatsu Nanozoomer 2.0-HT slide scanners have different resolution and are used to produce RGB high-power fields (HPFs). Annotations for the image coordinates of each mitosis are made by two senior pathologists, where if one disagrees a third will give the final say. The X40 resolution training set provided consists of 1,136 frames containing a total of 749 labeled mitotic cells. Aperio images are sized 1539 $\times$ 1376 pixels, and Hamamatsu images are 1663 $\times$ 1485 pixels.

Since the time of the contests both have been very commonly re-used among the research in this area thus far. Therefore, testing with these datasets will help compare the results to other published works. 

\paragraph{Preprocessing}
The annotations of \textit{official} test set for the ICPR contests is unavailable to the public so part of this training set was used for the test set. This is similar to what most other research groups (such as those referenced) have done. Specifically, the test set was selected randomly by extracting a set of images containing approximately the average number of mitosis in one HPF slide from the training set. The training set was required to be artificially augmented to increase the density of mitosis due to thier low concentration. Image patches in mitotic regions are first cropped into approximately 64 equal sized subsections from each HPF after being converted to JPEG. Then, the bounding box coordinates were then created by adding 20 pixels in the upper left and lower right directions from the derived annotated centroid coordinates. Finally, each image and its coordinates are expanded by 1.7 times, as this is the appropriate input image size and scale for the network setup.

\subsubsection{Accuracy Calculation}

The score in the same way as the contestants, using the F-score. According to the contest evaluation criteria, a correct detection (true positive) is the one that lies within 32 pixels from this centroid of the ground truth mitosis. The F-score is a harmonic mean of precision and recall (sensitivity), as described below. 

\begin{equation}  F-score = \frac{2 \cdot (precision \cdot sensitivity) }{(precision + sensitivity)} \end{equation} 


\subsection{Software \& Hardware}
The Multi-stream model was developed based on a modified version of the Faster RCNN \cite{renNIPS15fasterrcnn} and evaluated with a Quadro 6000 Cloud GPU.

\section{Results}

The resulting F-score that was obtained is 0.508. To start, this is significantly higher than contest winners score of 0.356. Table \ref{tab1} also shows a comparison of the precision, recall, and F-score to other top-performing groups who have more recently evaluated this type of deep learning network on the different ICPR 2014 MITOSIS training data subsets. The top results described in literature either used a version of the regional CNN (eg. RCNN) or a deep cascaded network. The Multi-stream Faster RCNN achieved a higher F-score than a single stream alone as well as most others who have tested their models on the public data. Further, this method takes only about 0.55s to test a single HPF, making it faster than other methods.

\begin{table}[htbp]
\caption{Performance Comparison on 2014 MITOSIS Dataset}
\begin{center}
\begin{tabular}{|c|c|c|c|}
\hline
\textbf{Method}&\multicolumn{3}{|c|}{\textbf{Results}} \\
\cline{2-4} 
  & \textbf{\textit{Precision}}& \textbf{\textit{Recall}}& \textbf{\textit{F-score}} \\
\hline

Lightweight RCNN \cite{li} & 0.40 & 0.45 & 0.427 \\
FRCNN & 0.43 & 0.44 & 0.435 \\
DeepMitosis (DeepDet+Seg+Ver) \cite{4} & 0.43 & 0.44 & 0.437 \\
CasNN \cite{chen} & 0.46 & 0.51 & 0.482 \\
MS-FRCNN & 0.48 & 0.54 & 0.508 \\
\hline
\end{tabular}
\label{tab1}
\end{center}
\end{table}

\section{Conclusion}

Mitotic count from HPF is a commonly used method to assess the level of progression of breast cancer, which is now the fourth most prevalent cancer. In this work, a segmented image map is generated with UNet and is input to a second stream of the Multi-stream Faster RCNN. Features from both streams are fused in the bilinear pooling layer. The additional detail features produced by UNet help to localize the mitotic regions and improve the accuracy over other methods and over a single stream alone. An F-score of 0.508 was obtained, which is higher than both the ICPR contest winners score and scores from other methods on the same data. Additionally, the technique is clinically applicable taking less than a second to test a single HPF.


\newpage


\bibliographystyle{IEEEtran}
\bibliography{root}
%

%
%

\end{document}